\documentclass[twocolumn]{article} 
\usepackage{lipsum} 
\usepackage[utf8]{inputenc} 
\usepackage[T1]{fontenc} 
\usepackage{graphicx} 
\usepackage{lipsum}         
\usepackage{natbib}         
\title{MV-GMN: State Space Model for Multi-View Action Recognition}

\author{
    Yuhui Lin\thanks{Equal contribution} \\ 
    Xi'an Jiaotong-Liverpool University 
    \and 
    Jiaxuan Lu\footnotemark[1] \\ 
    Shanghai Artificial Intelligence Laboratory 
    \and 
    Jiahao Zhang \\ 
    Xi'an Jiaotong-Liverpool University
}

\date{} 
\begin{document}

\maketitle

\begin{abstract}
Recent advancements in multi-view action recognition have largely relied on Transformer-based models. While effective and adaptable, these models often require substantial computational resources, especially in scenarios with multiple views and multiple temporal sequences. Addressing this limitation, this paper introduces the MV-GMN model, a state-space model specifically designed to efficiently aggregate multi-modal data (RGB and skeleton), multi-view perspectives, and multi-temporal information for action recognition with reduced computational complexity. The MV-GMN model employs an innovative Multi-View Graph Mamba network comprising a series of MV-GMN blocks. Each block includes a proposed Bidirectional State Space Block and a GCN module. The Bidirectional State Space Block introduces four scanning strategies, including view-prioritized and time-prioritized approaches. The GCN module leverages rule-based and KNN-based methods to construct the graph network, effectively integrating features from different viewpoints and temporal instances. Demonstrating its efficacy, MV-GMN outperforms the state-of-the-arts on several datasets, achieving notable accuracies of 97.3\% and 96.7\% on the NTU RGB+D 120 dataset in cross-subject and cross-view scenarios, respectively. MV-GMN also surpasses Transformer-based baselines while requiring only linear inference complexity, underscoring the model's ability to reduce computational load and enhance the scalability and applicability of multi-view action recognition technologies.
\end{abstract}

%
\section{Introduction}
\begin{figure}[t]  
	\centering  
	\includegraphics[width=0.5\textwidth]{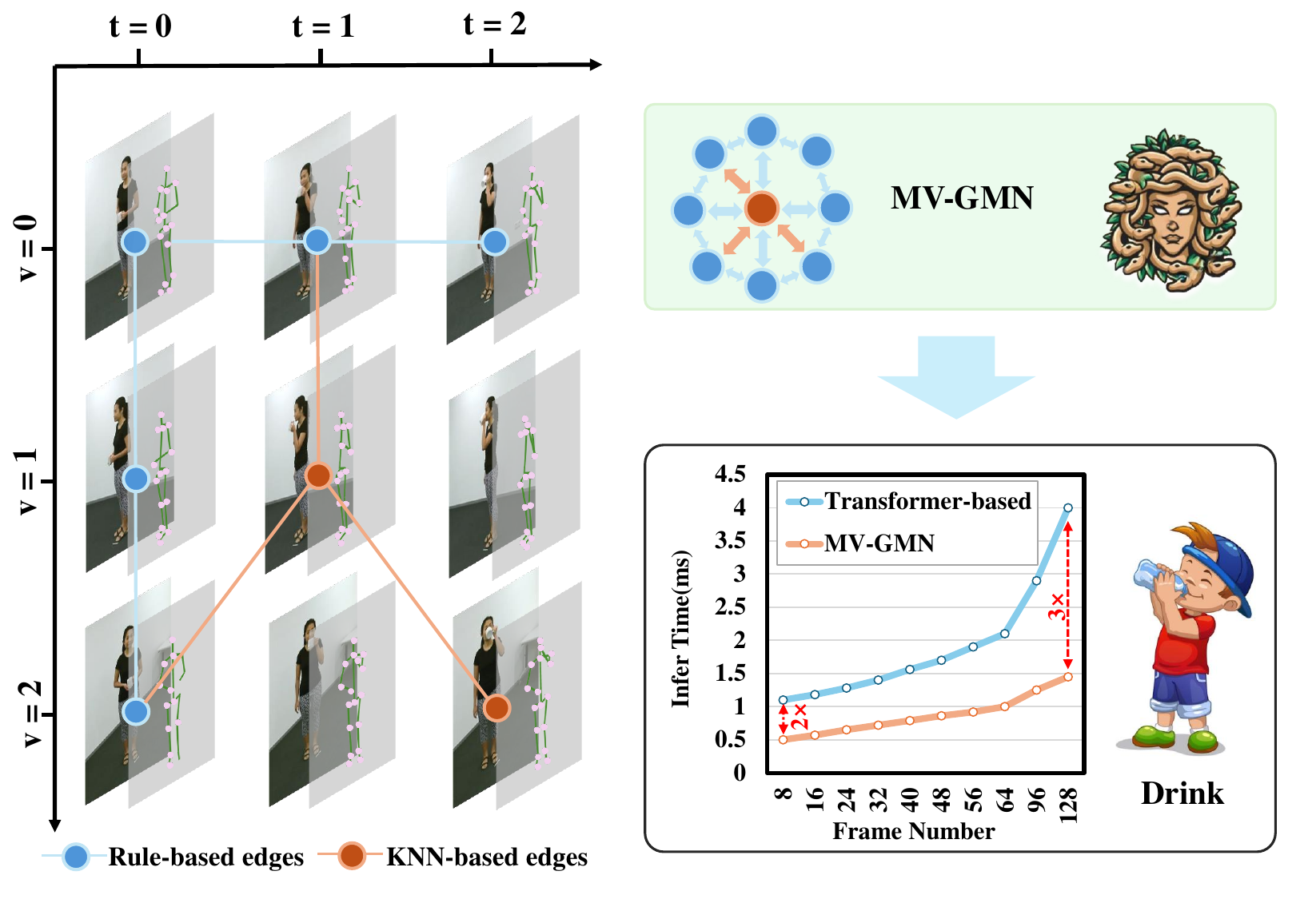} 
    \caption{\textbf{Overview of the framework}. The proposed MV-GMN employs rule-based and KNN-based edges to correlate temporal and viewpoint information, which is then fed into the proposed Multi-View Graph Mamba network. MV-GMN outperforms Transformer-based models while only requiring linear inference complexity for multi-modal, multi-view, multi-temporal sequence action recognition.
} 
	\label{fig:intro} 
\end{figure}
Action recognition is a pivotal task in computer vision with wide applications in surveillance \cite{elharrouss2021combined}, human-computer interaction \cite{lv2022deep}, and healthcare \cite{zhou2020deep,gao2023action}. It can replace human expertise in fields like sports training and physical therapy, where it provides insights into performance and recovery assessments \cite{host2022overview}.

Existing research primarily focuses on action recognition in the realm of single or multi-modal domains, particularly involving RGB video and skeleton data. RGB frames offer detailed visual information but is sensitive to background and lighting conditions, whereas skeletal data exhibits greater robustness \cite{wang20203dv, wang2018temporal}. Techniques such as Temporal Segment Networks (TSN) and 3D CNNs have been utilized to enhance visual temporal modeling in the RGB videos \cite{carreira2017quo}, while RNNs \cite{li2018independently}, CNNs \cite{zhang2018fusing}, and GCNs have been applied to the skeletal data \cite{ caetano2019skelemotion}. Single RGB videos often lack spatial information about individuals, which can be compensated by skeletal data. Therefore, researchers have explored fusion strategies\cite{bruce2022mmnet, lu2023exploring, zolfaghari2017chained}, such as Zhao et al.'s dual-stream framework for evaluating feature and score fusion \cite{zhao2017two}, Song et al.'s skeletal-guided feature extraction method \cite{song2018skeleton}, and the incorporation of attention mechanisms to enhance joint modeling of these modalities \cite{das2020vpn}.


In terms of viewpoints, most methods primarily rely on single-view processes, which often leads to difficulties in comprehensively understanding spatial information of individuals, especially in occluded scenes. Therefore, researchers have attempted to aggregate information from different viewpoints\cite{kim2023cross, bian2023global, gao2024hypergraph}, often leveraging attention mechanisms\cite{shah2023multi, siddiqui2024dvanet}. However, Transformer-based methods typically face high computational demands, particularly when modeling multi-view long-sequence videos. 

\begin{figure*}[htb]  
	\centering  
	\includegraphics[width=\textwidth]{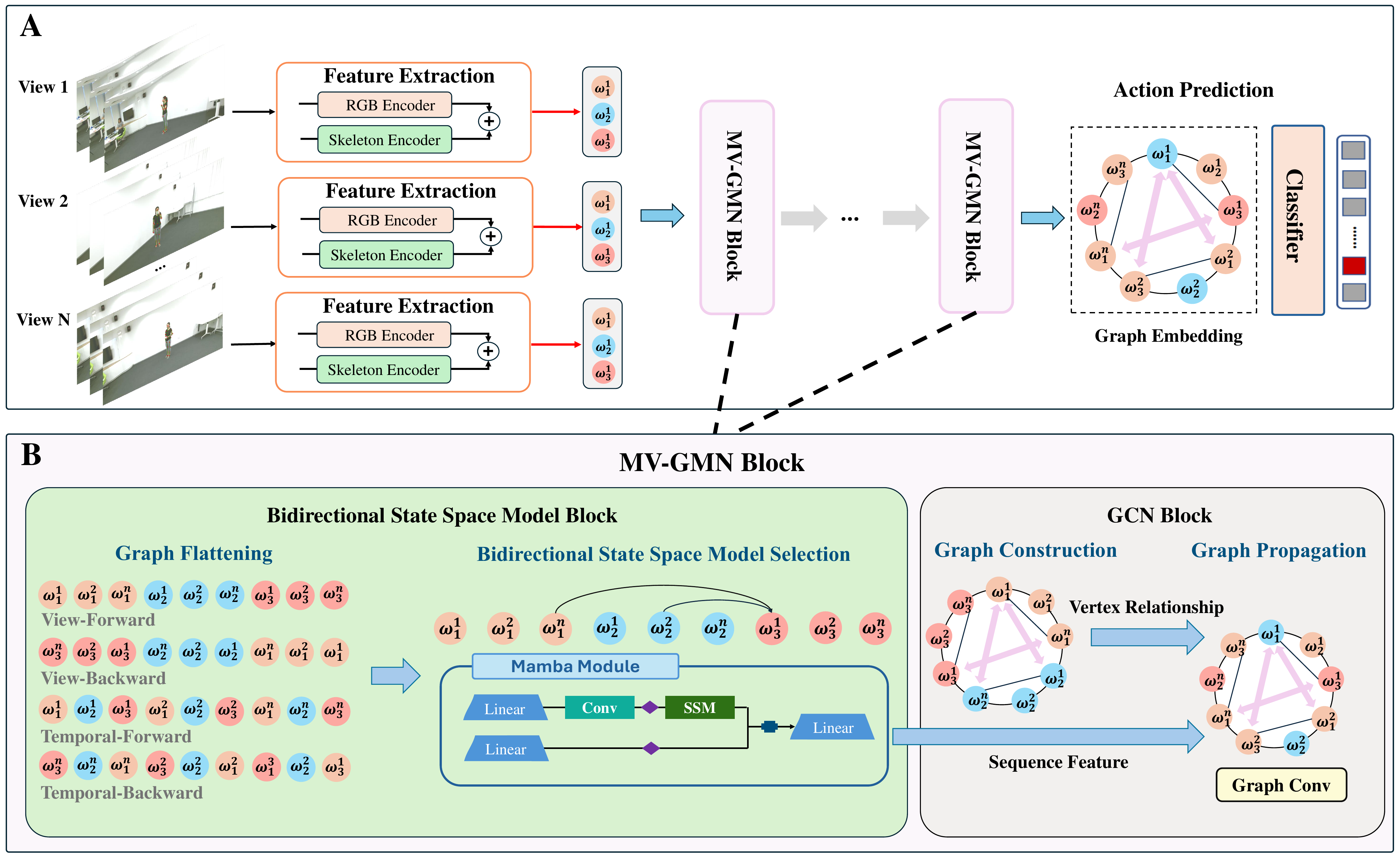} 
	\caption{\textbf{Overview of the MV-GMN architecture.} A) The framework's overall workflow includes Multi-Modal Feature Extraction, Multi-View Graph Mamba Network, and Action Prediction. B) Each MV-GMN Block consists of a Bidirectional State Space Model Block and a GCN block.} 
	\label{fig:pipeline} 
\end{figure*}

Recently, SSM-based methods like Mamba have demonstrated notable results in sequence processing \cite{gu2023mamba, patro2024simba, li2024videomamba} and multi-modal fusion \cite{dong2024fusion, li2024coupled, li2024mambadfuse} through linear inference complexity. However, there are currently no viewpoint fusion methods based on SSMs. To address the challenges in multi-view action recognition, we propose a state space model (SSM) based on graph convolutional networks, named \textbf{MV-GMN}, designed to efficiently model multi-view and multi-temporal sequences, as depicted in Figure~\ref{fig:pipeline}. 
As shown in Figure~\ref{fig:intro}, the framework effectively integrates features from various viewpoints and temporal instances, leveraging the graph's ability to model view-temporal correlations, while employing SSM's ability to efficiently process features. Specifically, initial RGB frames and skeletons are fed into a RGB encoder and a skeleton encoder, respectively. These features are then fed into the Multi-View Graph Mamba network including a series of MV-GMN blocks for feature aggregation. Each MV-GMN block comprises a Bidirectional State Space Block and a GCN block. The Bidirectional State Space Block introduces four scanning strategies, encompassing view-prioritized and time-prioritized approaches. Meanwhile, the GCN module employs rule-based and KNN methods to construct the graph structure, effectively integrating features from various viewpoints and temporal instances, which can capture both explicit and implicit relationships among features. Graph convolution layers are employed to obtain the graph embedding for the final action recognition. Extensive experiments in cross-subject and cross-view scenarios demonstrate superior improvements over previous methods.

Overall, the primary contributions of this paper can be summarized as follows: 
\begin{itemize}
	\item We are the pioneers in utilizing SSM for multi-modal, multi-view, multi-temporal sequence action recognition, marking the first attempt of its kind to date.
	\item We introduce a framework called MV-GMN, which employs a Multi-View Graph Mamba for action recognition, effectively integrating features across different viewpoints and temporal segments.
	\item Through experiments in cross-subject and cross-view scenarios, the proposed MV-GMN achieves significant improvements in multi-view action recognition compared to SOTA methods.
\end{itemize}

\section{Related Work}
\subsection*{Multi-View Action Recognition}
Recent studies in action recognition have increasingly addressed the complexities of multi-view action recognition, recognizing its significance in both practical scenarios and the field of representation learning \cite{kong2017deeply, bian2023global}. Predominant methods in this domain typically utilize skeletal, RGB, or multi-modal data, each presenting its own set of challenges. 

RGB-based methods improve action recognition accuracy and robustness through advanced feature learning tailored to varying viewpoints \cite{shah2023multi, siddiqui2024dvanet}.For skeleton-based action recognition, Graph Convolutional Networks (GCNs) are widely used \cite{chen2021channel}, with ST-GCN \cite{yan2018spatial} as a key model enhanced by multiscale and self-attention techniques \cite{li2019spatio, li2019actional}. In contrast, 2D-CNN methods transform skeleton sequences into pseudo-images \cite{asghari2020dynamic}. Few studies use 3D-CNNs \cite{hernandez20173d}.

Building on RGB and skeleton multi-modal fusion, \cite{zhao2017two} introduced a dual-stream framework using CNN and RNN, while \cite{baradel2017human, zolfaghari2017chained} developed three-stream 3D-CNN networks for modeling motion and images \cite{zhao2017two}. However, recent approaches like those by \cite{das2019toyota} emphasize feature fusion at the model level, employing cross-modal attention mechanisms to enhance recognition accuracy \cite{das2020vpn, bruce2022mmnet}

Simultaneously, many researchers have attempted to fuse viewpoints based on attention mechanisms, such as 3D Deformable Attention \cite{kim2023cross}, and through contrastive learning approaches \cite{bian2023global}. However, some Transformer-based methods often face high computational resource demands, especially when involving long-sequence video modeling. 

\subsection*{State Space Model}
Mamba is a neural network based on state space models, which improves the efficiency of sequential data processing through diagonal structured matrices \cite{gu2023mamba} or scalar multiplication by identity matrices \cite{dao2024transformers}. Simba optimizes the convergence issues specific to the Mamba model \cite{patro2024simba}. Additionally, by integrating with graph neural network technology, Mamba effectively processes graph data through bidirectional state space model architectures and input-dependent node selection mechanisms \cite{behrouz2024graph, wang2024graph}. Furthermore, applications of Mamba in the visual domain, such as Vision Mamba \cite{zhu2024vision}, VMamba \cite{liu2024vmamba}, demonstrate its efficiency in processing while maintaining accuracy. 

The State Space Model has not only shown excellent performance in sequence modeling but also demonstrated commendable capabilities in multimodal fusion \cite{dong2024fusion, li2024coupled, li2024mambadfuse}. Despite these achievements, the State Space Model has not yet been applied to multi-view and multi-sequence fusion for action recognition. 


\section{Methodology}
In this section, we present our proposed framework for multi-modal, multi-view, multi-temporal sequence action recognition. First, we introduce our feature extraction methods for RGB frames and skeletons and the fusion of RGB frames and skeletons. Following that, we focus on the proposed Bidirectional State Space Model Block within the MV-GMN Block. Next, We elaborate on our graph construction methods, the vertex propagation mechanism and the classification method.

\subsection*{Multi-Modal Feature Extraction}
In this section, we discuss methods for extracting features from individual RGB frames and skeletons, as well as the fusion of RGB frames and skeletons.

For the feature extraction of RGB frames, we first perform image cropping, using skeletal data to extract the foreground from the images. Next, we utilize the DeiT\cite{touvron2021training} as the RGB encoder, which processes the input image by dividing it into a fixed number of patches of fixed size. Each patch, sized \(t \times h \times w\), is encoded through a linear projection \(L\), transforming them into a sequence for model input. The process includes a specific token \(z^{\mathit{rgb}}_{t}\), which integrates all patches and positional embeddings \(p\) to preserve spatial information. The RGB features are extracted from video frames as follows:
\begin{equation}
z_{\mathit{rgb}} = \left[z^{\mathit{rgb}}_{1}, z^{\mathit{rgb}}_{2}, \ldots, z^{\mathit{rgb}}_{t}\right]
\end{equation}

For the feature extraction from skeletons, we have modified the Sparse-MSSTNet\cite{cheng2024dense} as the skeleton encoder, employing a multi-scale convolution kernel approach, to focus more on single-frame spatial two-dimensional feature extraction and reduce the influence of temporal fusion. The adaptation allows for the input of $t$ frames of skeleton to generate $t$ skeletal frame tokens \(z^{\mathit{sk}}_{t}\), and retrieves the necessary skeletal tokens \(z_{\mathit{sk}}\) for the video:
\begin{equation}
z_{\mathit{sk}} = \left[z^{\mathit{sk}}_{1}, z^{\mathit{sk}}_{2}, \ldots, z^{\mathit{sk}}_{t}\right]
\end{equation}

In terms of modality fusion, we first freeze the RGB encoder and the skeleton encoder. Then, we adopt cross-attention to allow the skeletal features to guide the attention mechanism towards relevant RGB features. In the design of the cross-attention module, specific trainable parameters are used to generate queries (Q), keys (K), and values (V). The skeletal features \( z_{\mathit{sk}} \) are utilized to generate the queries:

\begin{equation}
    Q = W_q z^{\mathit{sk}}_{t}
\end{equation}

The RGB features \( z^{\mathit{rgb}}_{t} \) are used to generate both keys and values:
\begin{equation}
    K = W_k z^{\mathit{rgb}}_{t}
\end{equation}
\begin{equation}
    V = W_v z^{\mathit{rgb}}_{t}
\end{equation}
where \( W_q \), \( W_k \), and \( W_v \) are the learnable weight matrices, tailored to modify the original features to produce representations suitable for the cross-attention mechanism.

The raw attention scores are computed by taking the dot product between the queries \( Q \) and keys \( K \), followed by normalization using the softmax function to obtain the final attention weights \( A \):

\begin{equation}
    A = \mathit{softmax}\left(\frac{QK^T}{\sqrt{d_k}}\right)
\end{equation}
where \( d_k \) is the dimensionality of the key vectors, used to scale the dot product to control the gradient flow. Finally, the weighted summation of values \( V \) using the attention weights \( A \) produces a comprehensive feature representation \( z_{\mathit{fused}} \) that integrates information from the skeletal and RGB features:

\begin{equation}
    z^{\mathit{fused}}_{t} = AV
\end{equation}
where \( z^{\mathit{fused}}_{t} \) represents the fusion of skeletal and RGB features.

Therefore, after performing feature fusion on a frame-by-frame basis, we integrate these features along the video timeline into a sequence feature \( Z_{fused} \), as shown in the following equation:
\begin{equation}
Z_{fused} = \left[z^{\mathit{fused}}_{1}, z^{\mathit{fused}}_{2}, \ldots, z^{\mathit{fused}}_{t}\right]
\end{equation}

\subsection*{Bidirectional State Space Model Block}

\begin{figure}[t]  
	\centering  
	\includegraphics[width=0.48\textwidth]{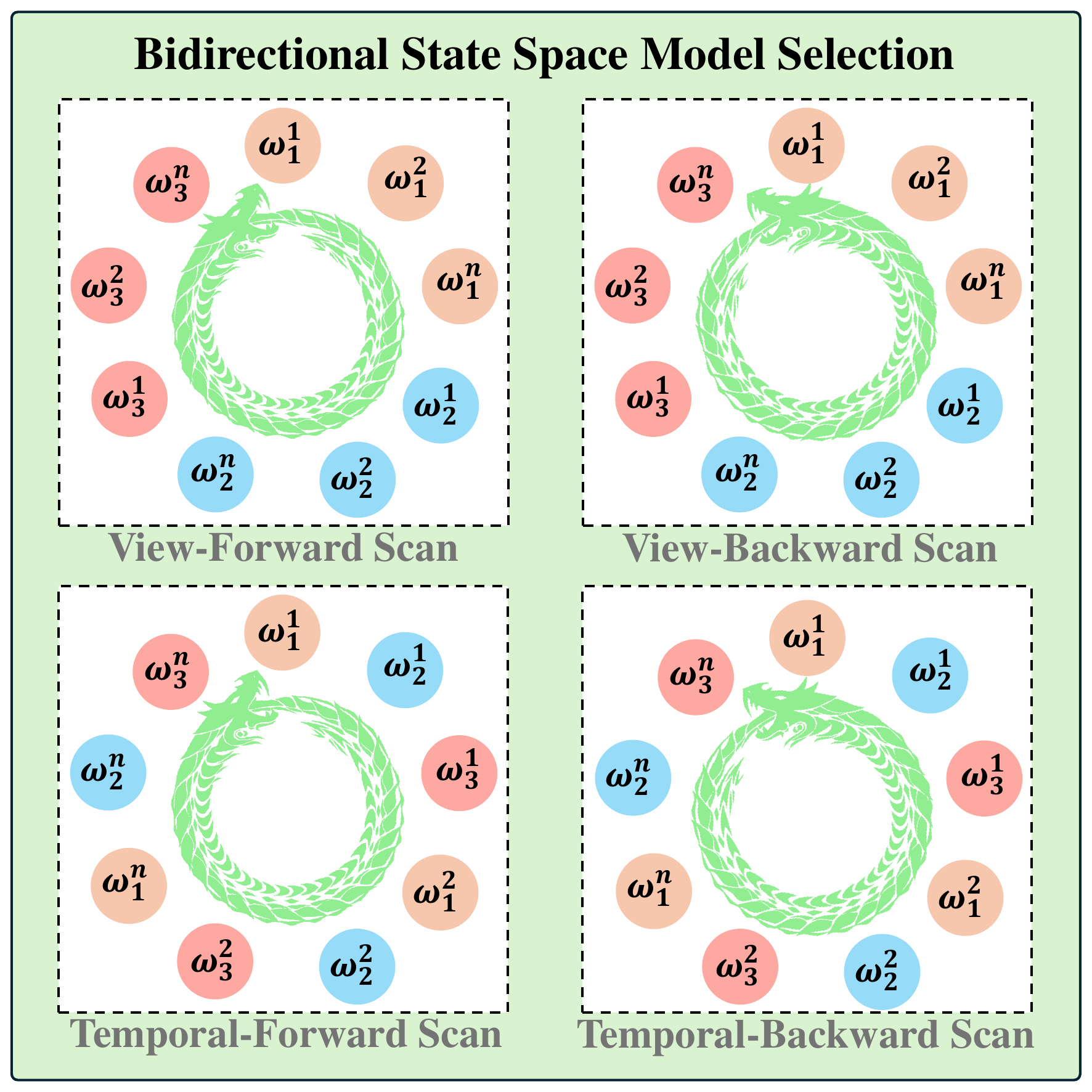} 
	\caption{Different scanning methods, including viewpoint-priority forward and backward scans, as well as temporal-priority forward and backward scans.} 
	\label{fig:bidirectionalSSMBlock} 
\end{figure}

In this section, we will elaborate on the Bidirectional State Space Model Block within the MV-GMN module, as shown in Figure~\ref{fig:bidirectionalSSMBlock}. Simultaneously, the RGB encoder, skeleton encoder, and cross-attention will be frozen.

We flatten the features fused from RGB frames and skeletons, resulting in four types of feature arrangements: the forward sequence prioritizing viewpoints \( P^{v}_{f} \), the backward sequence prioritizing viewpoints \( P^{v}_{b} \), the forward sequence prioritizing time \( P^{t}_{f} \), and the backward sequence prioritizing time \( P^{t}_{b} \), as demonstrated in the following equation:

\begin{equation}
P^{v}_{f} = \left[w^{\mathit{1}}_{1}, w^{\mathit{2}}_{1}, \ldots, w^{\mathit{v}}_{t}\right]
\end{equation}
\begin{equation}
P^{v}_{b} = \left[w^{\mathit{v}}_{t}, w^{\mathit{v-1}}_{t}, \ldots, w^{\mathit{1}}_{1}\right]
\end{equation}
\begin{equation}
P^{t}_{f} = \left[w^{\mathit{1}}_{1}, w^{\mathit{1}}_{2}, \ldots, w^{\mathit{v}}_{t}\right]
\end{equation}
\begin{equation}
P^{t}_{b} = \left[w^{\mathit{v}}_{t}, w^{\mathit{v}}_{t-1}, \ldots, w^{\mathit{1}}_{1}\right]
\end{equation}
where \( w^{\mathit{v}}_{t} \) denotes the features from different viewpoints, which originate from \( Z_{\mathit{fused}} \).

After flattening, we apply a one-dimensional convolution to each of the four sequences, as shown in the following equation:
\begin{equation}
    P' = \mathit{ReLU}(\mathit{Conv1D}(P, W_c))
\end{equation}
Here, \( P \) represents the set of the four flattening methods, \( P' \) denotes the result of convolving \( P \) with \( W_c \), and \( W_c \) represents the weights of the one-dimensional convolution layer. 

Concurrently, we employ Mamba based on the State Space Model (SSM) to model the four sequences within \( P \), as depicted in Figure~\ref{fig:bidirectionalSSMBlock}. Each sequence passes through a Mamba layer, sequentially comprising a linear layer, a convolutional layer, and an SSM. Moreover, the output from the SSM undergoes a residual connection with the original sequence, as demonstrated in the following equation:
\begin{equation}
    P_{Conv} = \mathit{Conv}(\mathit{Linear}(P'))
\end{equation}
\begin{equation}
    P_{SSM} = \mathit{SSM}(P_{Conv})
\end{equation}
\begin{equation}
    X = \mathit{Linear}(P_{SSM}+\mathit{Linear}(P'))
\end{equation}
where \( X \) denotes the result of each sequence in \( Mamba \) having been processed through the selected State Space Model.

Specifically, we initially perform state space modeling with a priority on the forward selection of viewpoints. Subsequently, we reorder it to prioritize the reverse selection of viewpoints. Following this, we further reorder the modeling to prioritize time.

\subsection*{Multi-View Graph Construction and Propagation}
\begin{figure}[h]  
	\centering  
	\includegraphics[width=0.4\textwidth]{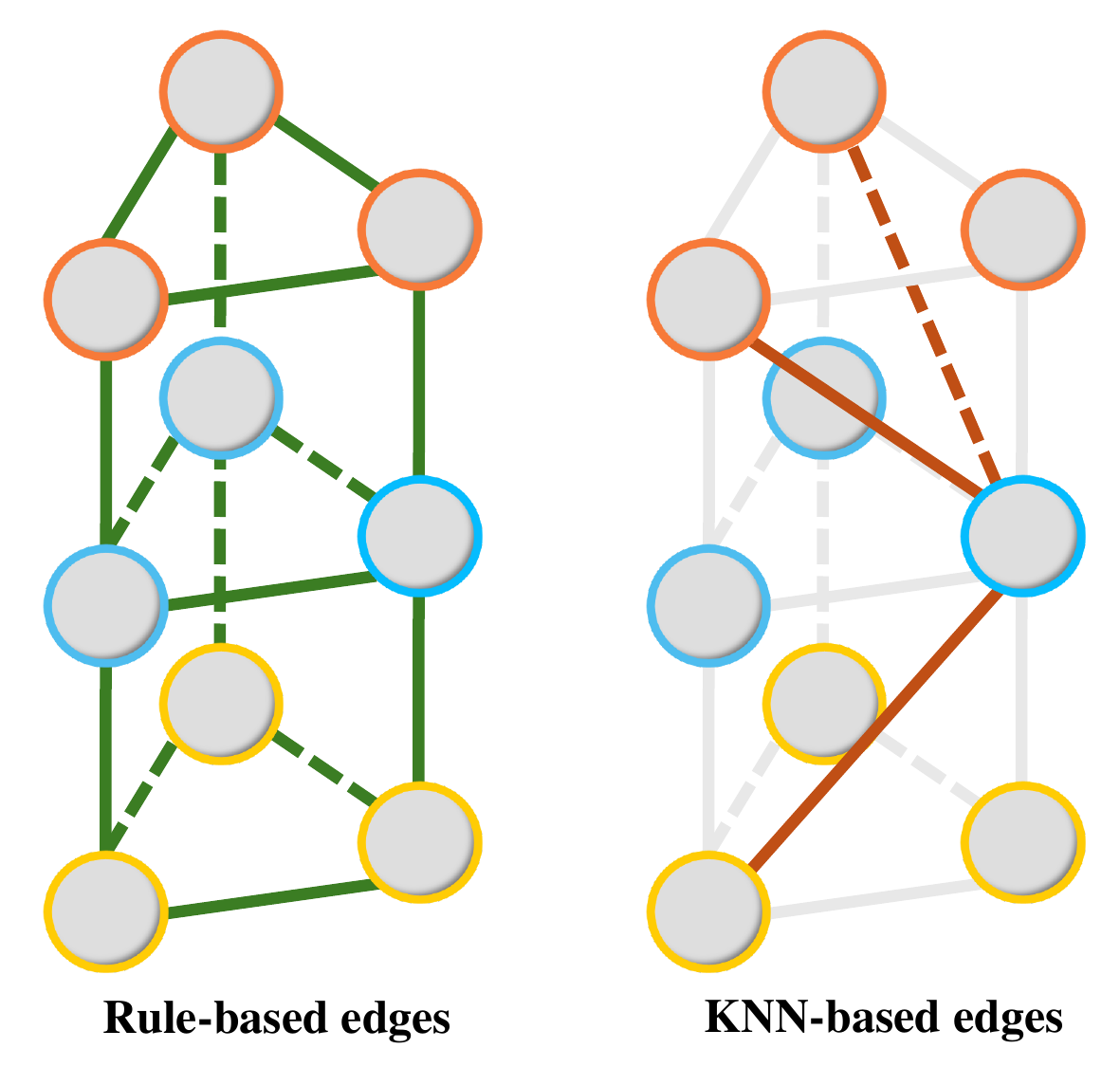} 
	\caption{The left shows the Rule-based edges based on different times and different viewpoints, while the right shows the KNN-based edges.} 
	\label{fig:knn} 
\end{figure}
Considering the challenges arising from the lack of information in a single viewpoint, the multi-view strategy for integrating features from diverse viewpoints and across different time intervals can greatly affect performance. In a multi-view setup, there exist sequential associations within the same viewpoint across various moments, as well as correlations among different viewpoints at the same instant. Thus, we propose a Multi-View Graph Mamba network to consolidate features spanning viewpoints and temporal dimensions, using rule-based and KNN-based edges to simulate both explicit and implicit connections, as shown in Figure~\ref{fig:knn}.

Specifically, we regard the one-dimensional features \( w_v^t \) under the viewpoint \( v \) and moment \( t \) as vertices, denoted as \( (v, t) \). In the multi-view scenario with \( V \) viewpoints and \( T \) time windows, there are a total of \( V \times T \) vertices. For the rule-based strategy, we employ two types of edges: the time-consistent edge \( \mathcal{E}_{\mathit{rule}}^{(t)} \) connects vertices of different moments within the same view, and the view-consistent edge \( \mathcal{E}_{\mathit{rule}}^{(v)} \) links vertices from various views at the same moment, which could be denoted as:
\begin{equation}
	\mathcal{E}_{\mathit{rule}}^{(t)} = \{ (v, t), (v', t') \mid v = v', t \neq t' \}
\end{equation}
\begin{equation}
	\mathcal{E}_{\mathit{rule}}^{(v)} = \{ (v, t), (v', t') \mid t = t', v \neq v' \}
\end{equation}

Subsequently, the rule-based edges are represented as \( \mathcal{E}_{\mathit{rule}} = \mathcal{E}_{\mathit{rule}}^{(t)} \cup \mathcal{E}_{\mathit{rule}}^{(v)} \). Regarding the KNN-based approach, for each vertex \( (v, t) \), we identify the \( k \) vertices within the embedding that show the greatest similarity, without considering their temporal or perspective alignment. Therefore, the set of KNN-based edges \( \mathcal{E}_{\mathit{knn}} \) is defined as:

\begin{equation}
	\mathcal{E}_{\mathit{knn}} = \{ (v, t), (v', t') \in N_k(v, t) \}
\end{equation}
where \( N_k(v, t) \) signifies the \( k \) vertices demonstrating the highest similarity to vertex \( (v, t) \) in terms of their embeddings. Subsequently, the two types of edge sets are combined to obtain the global edge set \( \mathcal{E} = \mathcal{E}_{\mathit{rule}} \cup \mathcal{E}_{\mathit{knn}} \). The graph utilizes the incidence matrix \( H \) to indicate whether the edges \( e \) contain the vertices \( (v, t) \), which can be expressed as:

\begin{equation}
	H((v, t), e) = \left\{
	\begin{array}{cl}
		1, &  (v, t) \in e \\
		0, &  (v, t) \notin e
	\end{array}
	\right.
\end{equation}

After the graph construction, we propagate features of the view-temporal vertices according to the general graph convolution method \cite{kipf2016semi}, as denoted: 
\begin{equation}
	X^{(l+1)} = \sigma \left( \tilde{D}^{-\frac{1}{2}} \tilde{A} \tilde{D}^{-\frac{1}{2}} X^{(l)} W^{(l)} \right)
\end{equation}
where \(X'^{(l)}\) represents the node features at the \(l\)-th layer, \(\tilde{A} = A + I\) is the adjacency matrix \(A\) with added self-connections \(I\), \(\tilde{D}\) is the degree matrix of \(\tilde{A}\), \(W^{(l)}\) is the weight matrix for the \(l\)-th layer, and \(\sigma\) denotes a nonlinear activation function.

Lastly, perform global average pooling separately on \( X \) and the original data \( P \). Specifically, calculate the average of all nodes across each feature dimension for both \( X \) and \( P \), resulting in two fixed-length vectors. These vectors are then concatenated to form a comprehensive feature vector. The vector is subsequently fed into a linear layer for the acrion classification.

\section{Experimental Results}
In this section, we conducted extensive experiments on the MV-GMN model. Initially, we compared MV-GMN with the latest multi-view action recognition models. Secondly, to verify whether MV-GMN can effectively perform multi-view modeling, we conducted analyses and evaluations on each component within MV-GMN. Finally, we compared the performance of MV-GMN under different parameters to prevent the model from overfitting or underfitting. Simultaneously, the principal objective of our experimentation was to demonstrate that the State Space Model can effectively model views and achieve favorable outcomes, all while utilizing a parameter count that is significantly lower than that of Transformer-based methods.
\subsection*{Datasets and Implementation Details}
\noindent \textbf{NTU RGB+D 60} includes 56,880 video samples from 40 participants, covering 60 action classes such as daily activities, health-related actions, and mutual actions. Data were captured from three different camera angles, providing depth maps, 3D skeletal, RGB, and infrared video data \cite{shahroudy2016ntu}.

\noindent \textbf{NTU RGB+D 120} extends the NTU RGB+D 60 with 114,480 samples and 120 action classes, enhancing sample and action diversity \cite{liu2019ntu}. 

\noindent \textbf{PKU-MMD} is a concise multi-modality benchmark for 3D human action understanding, comprising over 2,000 video sequences, nearly 20,000 action instances, and 5.4 million frames. It records using Kinect v2 sensors and includes data types like RGB, depth maps, infrared, and skeletal data \cite{liu2017pku}.

\noindent \textbf{Implementation Details} In all experiments, RGB videos are divided into 8 segments, while Skeleton videos are divided into 16 segments, with one frame randomly sampled from each segment on average. For RGB feature extraction, we utilize a fixed-parameter Deit-B. For Skeleton data extraction, we use an improved Sparse-MSSTNet. Additionally, we employ 4 MV-GMA blocks with an SSM state expansion factor of 64. For optimization, we use the SGD algorithm with an initial learning rate of 0.0025, reducing the learning rate by a factor of 0.1 if there is no improvement in the model within 5 epochs. The batch size is set to 64. At the same time, the results of our experiment running 64 epochs under the NVIDIA 4090. Furthermore, when comparing with the state-of-the-art (SOTA), we used Top-1 accuracy as the comparison metric.

\subsection*{Comparison with Existing Methods}

Our proposed MV-GMN model will be compared with several multi-view action recognition works that integrate multi-modal information from RGB and Skeleton data. The results show that MV-GMN outperforms the state-of-the-art methods in both cross-subject and cross-view scenarios.

\begin{table}[t]
	\centering
	\newcommand{\thickhline}{\noalign{\hrule height 1.2pt}}
	\begin{tabular}{l|c|c}
		\thickhline
		\textbf{Model} & \textbf{CS} & \textbf{CV} \\ \hline
		\cite{liu2018recognizing} & 91.7 & 95.2 \\
		\cite{das2020vpn}  & 95.5 & 98.0 \\ 
		\cite{davoodikakhki2020hierarchical} & 95.7 & 98.8 \\
		\cite{duan2022revisiting} & 97.0 & 99.6 \\
		\cite{bruce2022mmnet} & 96.0 & 98.8 \\
		\cite{shah2023multi} & 93.7 & 98.9 \\
		\cite{reilly2024just} & 96.3 & 99.0 \\
		\cite{cheng2024dense} & 97.4 & 99.4 \\
		\hline
            \textbf{MV-GMN(View-Prioritized)} & \textbf{97.4} & \textbf{99.1} \\
            \textbf{MV-GMN(Time-Prioritized)} & \textbf{96.1} & \textbf{98.3} \\
		\textbf{MV-GMN(View-Time)} & \textbf{98.2} & \textbf{99.7} \\
		\thickhline
	\end{tabular}
	\caption{Comparison between MV-GMN and previous SOTA multi-modal models on NTU RGB+D 60}
	\label{tab:NTURGBD60}
\end{table}
\begin{table}[t]
	\centering
	\newcommand{\thickhline}{\noalign{\hrule height 1.2pt}}
	\begin{tabular}{l|c|c}
		\thickhline
		\textbf{Model} & \textbf{CS} & \textbf{CV} \\ \hline
		\cite{liu2018recognizing} & 64.6 & 66.9 \\
		\cite{das2021vpn++} & 90.7 & 92.5 \\
		\cite{duan2022revisiting} & 96.4 & 95.3 \\
		\cite{bruce2022mmnet} & 94.4 & 92.9 \\
		\cite{ahn2023star}	& 92.8 & 91.1 \\
		\cite{reilly2024just} & 96.1 & 95.1 \\
		\cite{liu2024explore} & 92.7 & 90.3 \\
		\cite{cheng2024dense} & 96.7 & 95.6 \\
		\hline
            \textbf{MV-GMN(View-Prioritized)} & \textbf{97.0} & \textbf{95.1} \\
            \textbf{MV-GMN(Time-Prioritized)} & \textbf{95.2} & \textbf{94.6} \\
		\textbf{MV-GMN(View-Time)} & \textbf{97.3} & \textbf{96.7} \\
		\thickhline
	\end{tabular}
	\caption{Comparison between MV-GMN and previous SOTA multi-modal models on NTU RGB+D 120}
	\label{tab:NTURGBD120}
\end{table}
\begin{table}[t]
	\centering
	\newcommand{\thickhline}{\noalign{\hrule height 1.2pt}}
	\begin{tabular}{l|c|c}
		\thickhline
		\textbf{Model} & \textbf{CS} & \textbf{CV} \\ \hline
		\cite{bruce2021multimodal} & 95.8 & 97.8 \\
		\cite{bruce2022mmnet} & 97.4 & 98.6 \\
		\cite{ahn2023star}	& 92.8 & 91.1 \\
		\cite{cheng2024dense} & 97.4 & 98.8 \\
		\hline
            \textbf{MV-GMN(View-Prioritized)} & \textbf{97.3} & \textbf{97.6} \\
            \textbf{MV-GMN(Time-Prioritized)} & \textbf{96.2} & \textbf{96.7} \\
		\textbf{MV-GMN(View-Time)} & \textbf{98.0} & \textbf{99.1} \\
		\thickhline
	\end{tabular}
	\caption{Comparison between MV-GMN and previous SOTA multi-modal models on PKUMMD}
	\label{tab:PKUMMD}
\end{table}

Table~\ref{tab:NTURGBD60} presents the performance of the MV-GMN model. Specifically, MV-GMN (View-Time) achieves a Top-1 accuracy of 98.2\% in the cross-subject scenario and 99.7\% in the cross-view scenario. These figures surpass the results of Cheng's model \cite{cheng2024dense} and Duan's model \cite{duan2022revisiting} by 0.8\% and 0.1\%, respectively. Additionally, MV-GMN (View-Prioritized) achieves accuracies of 97.4\% and 99.1\% in the cross-subject and cross-view scenarios, respectively. MV-GMN (Time-Prioritized) achieves accuracies of 96.1\% and 98.3\% in the cross-subject and cross-view scenarios, respectively.

Additionally, as shown in Table~\ref{tab:NTURGBD120}, the MV-GMN (View-Time) model demonstrates impressive performance on the NTU RGB+D 120. In the cross-subject and cross-view scenarios, it achieves Top-1 accuracies of 97.3\% and 96.4\%, respectively. These results are notably higher by 0.6\% and 0.7\% compared to the model proposed by Cheng \cite{cheng2024dense}. Furthermore, MV-GMN (View-Prioritized) achieves accuracies of 97.0\% and 95.1\% in the cross-subject and cross-view scenarios, respectively. Meanwhile, MV-GMN (Time-Prioritized) achieves accuracies of 95.2\% and 94.6\% in the cross-subject and cross-view scenarios, respectively.

Furthermore, as shown in Table~\ref{tab:PKUMMD}, our MV-GMN (View-Time) model exhibits outstanding performance on the PKUMMD. In the cross-subject and cross-view scenarios, it achieves Top-1 accuracies of 98.0\% and 99.1\%, respectively. These results exceed those of Cheng's model \cite{cheng2024dense} by 0.6\% and 0.3\%, respectively. Furthermore, MV-GMN (View-Prioritized) achieves accuracies of 97.3\% and 97.6\% in the cross-subject and cross-view scenarios, respectively. Meanwhile, MV-GMN (Time-Prioritized) achieves accuracies of 96.2\% and 96.7\% in the cross-subject and cross-view scenarios, respectively.

On the three datasets, it can be observed that the performance of MV-GMN (View-Prioritized) consistently surpasses that of MV-GMN (Time-Prioritized). Additionally, as illustrated in Figure 1, within MV-GMN (View-Prioritized), the three viewpoints at the same time point are closely spaced in the sequence, and the consideration of temporal factors is confined to either forward or reverse order. In contrast, in MV-GMN (Time-Prioritized), the three viewpoints at the same time point are more widely spaced in the sequence, and the consideration of temporal factors involves more frequent forward or reverse scanning, increasing the complexity of model construction.

\subsection*{Abaltion Studies}
In this section, we will analyze the effectiveness of the MV-GMN(View-Time) model in fusing RGB and Skeleton data, as well as its performance in multi-view temporal and spatial fusion on the NTU RGB+D 120.

\begin{table}[t]
	\centering
	\newcommand{\thickhline}{\noalign{\hrule height 1.2pt}}
	\begin{tabular}{l|c|c}
		\thickhline
		\textbf{Model} & \textbf{CS} & \textbf{CV} \\ \hline
		Deit-S & 87.7 & 89.1 \\
		Sparse-MSSTNet & 82.7 & 84.3 \\
		Mean Fusion & 94.9 & 95.6 \\
		Linear Fusion & 95.1 & 95.8 \\
		\hline
		\textbf{Cross-attention fusion} & \textbf{97.3} & \textbf{96.7} \\
		\thickhline
	\end{tabular}
	\caption{Fusion of RGB and skeleton Performance Comparison}
	\label{tab:efffuse}
\end{table}

\begin{table}[h]
	\centering
	\newcommand{\thickhline}{\noalign{\hrule height 1.2pt}}
	\begin{tabular}{l|c|c|c|c}
		\thickhline
		\textbf{Model} & \textbf{CS} & \textbf{CV} & \textbf{Params} & \textbf{Time (ms)}\\ \hline
            Linear & 94.1 & 94.8 & 20.92M & 29.53\\
            Transformer & 96.0 & 95.7 & 30.81M & 29.99\\
            Mamba & 96.3 & 95.9 & 27.89M & 29.84\\
            GCN & 94.8 & 95.1 & 22.10M & 30.00\\
            KNN-GCN & 96.3 & 95.9 & 22.10M & 30.02\\
            T-KNN-GCN & 96.9 & 96.4 & 31.99M & 30.06\\
		\hline
		\textbf{MV-GMN} & \textbf{97.3} & \textbf{96.7} & 29.08M & 29.88\\
		\thickhline
	\end{tabular}
	\caption{Fusion of view and time Performance Comparison}
	\label{tab:stab}
\end{table}

\paragraph{Effectiveness of RGB and Skeleton Fusion}
The efficiency of fusing RGB and skeleton data using a cross-attention mechanism is analyzed. As illustrated in Table~\ref{tab:efffuse}, we compared the performance of averaging and linear fusion methods. The results reveal that the averaging method outperforms the linear approach by 2.2\% and 0.9\% in cross-subject scenarios, respectively.

\paragraph{Effectiveness of Temporal and Viewpoint Fusion}
In our discussion on integrating viewpoint and temporal data, we utilized various methods including linear, transformer,mamba ,GCN, kNN-GCN,and T-KNN-GCN. We attempted to replace the position of MV-GMN in the network with linear, transformer,mamba ,GCN, kNN-GCN,and T-KNN-GCT to evaluate the effectiveness of different fusion strategies, as shown in Table~\ref{tab:stab}. 

Firstly, in our experiments with Mamba and Transformer, we found that Mamba surpasses the Transformer in multi-view and multi-temporal modeling capabilities while having 2.29M fewer parameters. Additionally, when propagating features using GCN, we incorporated linear layers which improved performance by 0.7\% and 0.3\% on cross-subject and cross-view tasks, respectively.Traditional edge construction methods, based solely on time and viewpoint, struggle to capture latent information.Therefore, we integrated a KNN approach for building edges (KNN-GCN), which resulted in enhancements of 1.5\% and 0.8\% in cross-subject and cross-view performance compared to GCN.

Furthermore, to further capture the hidden relationships between feature points, we employed Transformer (T-KNN-GCN) and Mamba (MV-GMN) based on KNN-GCN. Among these, MV-GMN outperformed KNN-GCN in cross-subject and cross-view tasks by 1.0\% and 0.8\%, respectively.Moreover, with 2.91M fewer parameters than T-KNN-GCN, MV-GMN not only exceeded its performance by 0.4\% and 0.3\% in cross-subject and cross-view tasks, respectively, but also achieved a faster inference time by 0.18ms on a single sample.

\subsection*{Hyperparameter Studies}
In this section, we conducted hyperparameter experiments with the MV-GMN (View-Time) model on the NTU RGB+D 120 to analyze the impact of the number of KNN neighbors and the number of MV-GMN Blocks on model performance. 

\begin{figure}[h]  
	\centering  
	\includegraphics[width=0.5\textwidth]{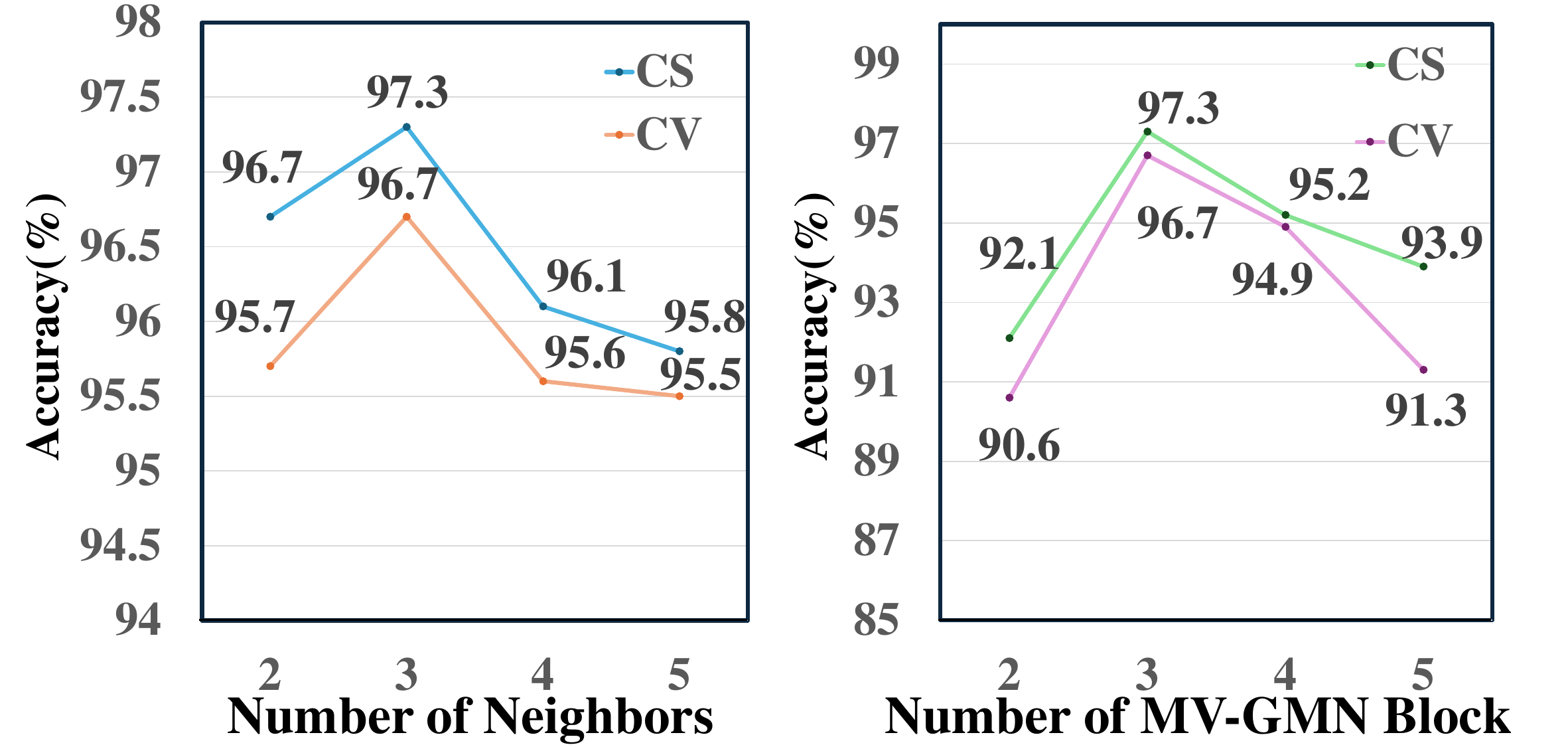} 
	\caption{The left shows the effects of different numbers of KNN neighbors, and the right shows the effects of different numbers of MV-GMN Blocks.} 
	\label{fig:hy} 
\end{figure}

\paragraph{Number of KNN Neighbors}
As shown in Figure~\ref{fig:hy}, the model performs best when the number of KNN (K-Nearest Neighbors) neighbors is set to 3. We believe that having too few neighbors may cause the MV-GMN model to focus excessively on details and overlook the overall trends of features, potentially leading to overfitting. Conversely, when the number of neighbors exceeds 3, it might introduce weakly related noise points into the MV-GMN modeling process, which could also lead to underfitting. These factors suggest that setting the neighbor count to 3 enables MV-GMN to effectively balance between capturing data details and generalizing, thus achieving optimal performance.

\paragraph{Number of MV-GMN Blocks}
As shown in Figure~\ref{fig:hy}, the model performs best when the number of MV-GMN Blocks is 4. The MV-GMN Block captures complex dependencies through the SSM and propagates directional relationships using graph neural networks. Therefore, when there are fewer than four MV-GMN Blocks, the model may struggle to fully capture the relationships between features. Conversely, if there are more than four MV-GMN Blocks, it could lead to excessive destruction of features and an increased risk of overfitting.

Furthermore, when the number of MV-GMN Blocks is 2, we first use a forward MV-GMN (View-Prioritized) Block, followed by a forward MV-GMN (Time-Prioritized) Block. As the number of Blocks increases to 4, 8, and 12, we respectively configure one, two, and three bidirectional MV-GMN (View-Time) Blocks.

\section*{Conclusion}
We propose a novel Multi-View Graph Mamba network for multi-modal, multi-view, multi-temporal sequence action recognition. The propsed MV-GMN integrates a Bidirectional State Space Block and a GCN block, enabling effective feature integration from various viewpoints and temporal instances while retaining linear inference complexity of state-space models. The proposed framework achieves SOTA performances on three large-scale multi-view action recognition datasets, underscoring its potential to enhance the scalability and applicability of action recognition technologies in resource-limited settings.

\bibliography{aaai25.bib}

\end{document}